\documentclass{new_tlp}

\usepackage{mathptmx}
\usepackage{url}
\usepackage{graphicx}
\usepackage{amsmath,amssymb,amsfonts}
\usepackage{xcolor}
\usepackage{booktabs}
\usepackage{listings}
\usepackage{enumitem}
\usepackage[linesnumbered,ruled,vlined]{algorithm2e}
 \SetAlFnt{\small} 
 \SetAlCapFnt{\small} 
 \SetAlCapNameFnt{\small} 
\usepackage[unicode,psdextra]{hyperref}
\usepackage{cleveref}
\usepackage{subcaption}
\usepackage{multirow}

\definecolor{myblue}{HTML}{1F77B4}

\SetCommentSty{mycommfont}

\newcommand{\dom}{\mathsf{dom}}
\newcommand{\var}{\mathsf{var}}

\newcommand{\enr}[1]{\textcolor{black}{#1}}

\newcommand{\set}[1]{\{#1\}}
\newcommand{\sval}{s^{val}}
\newcommand{\ssup}{s^{sup}}

\title[GPU-accelerated Table Constraint]
{GPU Accelerated Compact-Table Propagation{\,}\thanks{Research partially
supported by Interdepartment Project on AI (Strategic Plan UniUD–22-25), 
and
by Unione europea-Next Generation EU, missione~4 componente~2, project MaPSART ``Future Artificial Intelligence (FAIR)'', PNRR.
A.Dovier and A.Formisano are members of GNCS-INdAM, Gruppo Nazionale per il Calcolo Scientifico.}}
\author[Santi, Tardivo, Dovier, Formisano]
{{Enrico} {Santi}\phantom{aaa}
 {Agostino} {Dovier}\phantom{aaa}
 {Andrea} {Formisano} \\
 University of Udine, DMIF, Udine, Italy\;
 \\ \email{santi.enrico@spes.uniud.it,\{agostino.dovier|andrea.formisano\}@uniud.it} \and
{Fabio Tardivo}\\
 New Mexico State University, Dept CS, Las Cruces, NM, USA\,
 \email{ftardivo@nmsu.edu}}

\begin{document}
\label{firstpage}
\maketitle

\begin{abstract}
Constraint Programming developed within Logic Programming in the Eighties; nowadays all Prolog systems encompass modules capable of handling constraint programming on finite domains demanding their solution to a constraint solver. This work focuses on a specific form of constraint, the so-called table constraint, used to specify conditions on the values of variables as an enumeration of alternative options. 
Since every condition on a set of finite domain variables can be ultimately expressed as a finite set of cases, Table can, in principle, simulate any other constraint. These characteristics make Table one of the most studied constraints ever, leading to a series of increasingly efficient propagation algorithms. 
Despite this, it is not uncommon to encounter real-world problems with hundreds or thousands of valid cases that are simply too many to be handled effectively with standard CPU-based approaches.
In this paper, we deal with the Compact-Table (CT) algorithm, the state-of-the-art propagation algorithms for Table. 
We describe how CT can be enhanced by exploiting the massive computational power offered by modern GPUs to handle large Table constraints. In particular, 
we report on the design and implementation of GPU-accelerated CT, on its integration into an existing constraint solver, and on an experimental validation performed on a significant set of instances.
\end{abstract}

\section{Introduction}
Table constraints play a crucial role in combinatorial problem solving, as they provide an explicit and efficient way to represent all allowed combinations of values for the variables they involve \cite{main}. Their flexibility makes them widely applicable in different domains such as scheduling, configuration, and in every context where constraints on variables must be explicitly enumerated.
Additionally, planning problems often found in constraint (logic) programming, can be conveniently formulated as Constraint Satisfaction Problems (CSP) over finite 
domains~(\citeNP{DBLP:journals/fuin/DovierFP10};
\citeyearNP{DBLP:conf/birthday/DovierFP11,DovierFP13}) where state transitions or action scheme can be expressed as table constraints.
The relevance of such a constraint has led to the development of several efficient algorithms (such as STR2, STR3, MDD4R, and Compact-Table) to enforce the generalized arc consistency (GAC) \cite{str2,str3,gac4}.

Following recent works in which parts of constraint solvering have been delegated to GPUs
(\citeNP{cumulative};\citeyearNP{DBLP:journals/logcom/TardivoDFMP23,DBLP:conf/cp/TardivoMP24}; \citeNP{TravasciCILC25}), with this paper we are pursuing this project going toward a more exhaustive parallel implementation of constraint logic programming~\cite{DBLP:journals/tplp/DovierFGHPR22}.

In this paper, we recall an efficient GAC algorithm for the table constraint, Compact-Table (CT) and we describe different implementations of GPU-accelerated propagators based on CT inserting them in
{\sc MiniCP}, a simple and extendable constraint solver
\cite{minicp}.

The paper is organized as follows:
Section~\ref{sec:bg} recalls some background on the Table constraint, the CT propagator and the  {\sc MiniCP} constraint solver. Section~\ref{sec:serial} describes an implementation of the serial propagator. 
Based on such an implementation, in Section~\ref{sec:parallel} different GPU accelerated propagators are described. Section~\ref{sec:experiments} reports on the experiments we run to validate the parallel implementations and discusses the obtained results.  In Section~\ref{sec:conclusion} we draw our conclusions.

Source code, tests instances, and results are available from   {\small\url{https://clp.dimi.uniud.it/sw}}.

\section{CSPs and the Table constraint} 
\label{sub:CT}
\label{sec:bg}

A CSP is a triple $P= \langle X, D, C \rangle$, where  $X$ is a finite set of variables, $D$ is set of (finite) domains for the variables, and $C$ is the set of \emph{constraints} over $X$.
For each $x \in X$ its domain is denoted by $\dom(x)$.
A constraint $c \in C$ involves a set of variables (its scope) $\var(c)=\{x_{1},\dots,x_{n}\} \subseteq X$ and induces a relation $rel(c) \subseteq \dom(x_1) \times \cdots \times \dom(x_n)$.

A \emph{solution} of $P$ is an assignment $\sigma$ of variables to values in their domains
that satisfies all the constraints in $C$, namely such that $\sigma(X) \in rel(c)$ for all $c \in C$.
A constraint $c$ such that $\var(c)=\{x_1,\dots ,x_n\}$ is GAC if and only if for all $x_i \in\var(c)$ and all $a \in \dom(x_i)$ there is a tuple $(a_1,\dots,a_{i-1},a,a_{i+1},\dots,a_{n}) \in rel(c)$.
In such case   we say the value $a$ is supported. 
Conversely, if $a$ is not supported, $a$ cannot belong to any solution and therefore it can be eliminated from $\dom(x_i)$. 
Constraint propagation algorithms iteratively removes unsupported values until the GAC is enforced (or unsatisfiability, namely not existence of a solution, is detected).

Several kinds of constraints on finite domain variables have been introduced and studied \cite{RossiHandbookCP}.
Our focus in this paper is on the {\sc table} (a.k.a., {\sc extensional}) constraint.
The table constraint can theoretically simulate all the other constraints since it explicitly lists the tuples of values of the relation~\cite{basicSmart}. 
\Cref{tbl:esempio:table} shows an example of a table constraint involving three variables $x_1,x_2,x_3$, all having $\{1,2,3,4\}$ as domain.
The constraint specifies all the admissible tuples $\tau_1,\ldots, \tau_5$ of values.

Compact-Table (CT) is an efficient algorithm to enforce GAC on table constraints \cite{main}. While its base version allows to express only positive and exhaustive (i.e., non-short) tables,\footnote{A Table $c$ is \emph{positive} if its tuples list the allowed values for $var(c)$. A table explicitly listing the disallowed tuples is said \emph{negative}. A table is \emph{short} if more than a one domain value can be specified in each cell.} extensions have been introduced by 
 \citeANP{basicSmart}\citeyear{basicSmart,negativeShort}. 
In what follows we describe serial and parallel implementations of CT. Although these implementations can also be used for short tables, in this paper we limit ourselves to the case of positive and exhaustive tables.

\begin{table}[tb]
{\small
    \centering
    \begin{subtable}[b]{0.18\textwidth}
        \centering
        \begin{tabular}{cccc}
                     & $x_1$ & $x_2$ & $x_3$ \\
            \midrule
            $\tau_1$ & 3 & 1 & 1 \\
            $\tau_2$ & 1 & 2 & 3 \\
            $\tau_3$ & 2 & 3 & 3 \\
            $\tau_4$ & 1 & 4 & 1 \\
            $\tau_5$ & 3 & 4 & 3 \\
        \end{tabular}
         \vspace*{5mm}
        \caption{}
        \label{tbl:esempio:table}
    \end{subtable}
~~~~~~~
\begin{subtable}[b]{0.38\textwidth}
        \centering
        \begin{tabular}{cccccc}
                & $\tau_1$ & $\tau_2$ & $\tau_3$  & $\tau_4$ & $\tau_5$ \\ 
                \midrule
                $[x_1,1]$ & 0 & 1 & 0 & 1 & 0 \\ 
                $[x_1,2]$ & 0 & 0 & 1 & 0 & 0 \\ 
                $[x_1,3]$ & 1 & 0 & 0 & 0 & 1\\
                $[x_1,4]$ & 0 & 0 & 0 & 0 & 0 \\ 
                $[x_2,1]$ & 1 & 0 & 0 & 0 & 0 \\
                $[x_2,2]$ & 0 & 1 & 0 & 0 & 0\\
                $\cdots$ & $\cdots$ & $\cdots$ & $\cdots$ & $\cdots$ & $\cdots$\\
                $[x_3,4]$ & 0 & 0 & 0 & 0 & 0\\
        \end{tabular}
        \caption{}
        \label{tbl:esempio:support}
    \end{subtable}
~~~~~~~
    \begin{subtable}[b]{0.26\textwidth}
        \centering
        \begin{tabular}{ccccc}
            $\tau_1$ & $\tau_2$ & $\tau_3$ & $\tau_4$ & $\tau_5$ \\
    	    \midrule
            0 & 1 & 0 & 1 & 0 \\
        \end{tabular}
        \vspace*{12mm}
        \caption{}
        \label{tbl:esempio:currtable}
    \end{subtable}
    \caption{A table constraint $c$ with 5 tuples on the variables $x_1,x_2,x_3$ with domain $\{1,2,3,4\}$ (a), the corresponding static $\mathit{supports}$ matrix (b), and a possible value of 
    {\it currTable} (c).}	
    \label{tbl:esempio}
    }
\end{table}

{\small
\begin{algorithm}[tb]
\small

    \SetKwInOut{Input}{input}
    \SetKwInOut{Output}{output}
    \Input{$c$ a table constraint, with $\var(c)=\{x_1,\dots,x_n\}$}
    $\sval\leftarrow \set{ i \mid |\dom(x_i)| \neq lastSizes[i]}$\;
    \lFor{$i \in  \sval$ }{
        $lastSizes[i] \leftarrow |dom(x_i)|$
    }
    $\ssup \leftarrow \set{i \mid |\dom(x_i)| > 1 }$\;
    $updateTable()$\;
    \lIf{currTable = $\vec{0}$}{
        $\mathit{fail()}$
    }
    $\mathit{filterDomains()}$\;
    \caption{enforceGAC()}
    \label{alg:enf}
\end{algorithm} 
}

{\small
\begin{algorithm}[tb]
\small
    \SetKwInOut{Input}{input}
    \SetKwInOut{Output}{output}
    \Input{{\it supports}, {\it currTable}, $\dom$,  $\Delta$, $\sval$}
    \For(\tcp*[f]{For all variables with domain reduced by the last step}) {$i \in \sval$}
    { 
        $mask \leftarrow \vec{0}$\tcp*[r]{Reset a mask of $t$ bit}
        \uIf(\tcp*[f]{Reset-based update is advantageous}) 
        {$|\Delta_{x_i}|<|\dom(x_i)|$}{
            \For(\tcp*[f]{Collect indexes of tuples to be removed}){$a \in \Delta_{x_i}$}{
                $mask \leftarrow mask \;|\; supports[x_i,a]$\;
            }
            $mask\leftarrow \,\,  {\sim}mask$\tcp*[r]{Complement the {\it mask}}
        }\Else
        (\tcp*[f]{Incremental update is preferable}){
            \For(\tcp*[f]{Collect indexes of supported tuples}){$a \in \dom(x_i)$}{
              $mask \leftarrow \,\, mask \;|\; supports[x_i,a]$\;
            }
        }
        $\mathit{currTable} \leftarrow \mathit{currTable} \; \& \; mask$\tcp*[r]{Update {\it currTable} applying the {\it mask}}
        \lIf{$\mathit{currTable} = \vec{0}$}{$break$}
    }    
    \caption{updateTable()}
    \label{alg:updateTable}
\end{algorithm}
}

Consider a table constraint $c$ expressed by a set of  $t$ rows (tuples) and $n = |\var(c)|$ columns.
Let $\tau_1,\dots,\tau_t$ be the tuples of $c$, and $\var(c) = \{x_1,\dots,x_n\}$. In CT, $c$ is represented by two data structures,
a Boolean matrix and a Boolean array, both stored as sequences of bits:
\begin{description}
    \item[$\mathit{supports}$] A static Boolean matrix of size $(|\dom(x_1)|+\cdots+|\dom(x_n)|) \times t$, for each variable $x_i$ and each value $v \in \dom(x_i)$, the cell $([v,a],j)$ is set to 1 iff $\tau_j[i] = v$.
    \item[$\mathit{currTable}$] A Boolean array of size $t$ indicating, for each tuple $\tau_j$, if $\bigwedge_{i=1}^{n} \tau_j[i] \in \dom(x_i)$. 
\end{description}
A tuple $\tau_i$ is \emph{valid} if and only if $\forall x_j \in var(c)$, $\tau_i[j] \in dom(x_j)$.
During the search process the domains of the variables are reduced. 
The bitset $\mathit{currTable}$ keeps track of domain reductions by setting the $i$-th bit to\;1 if the $i$-th tuple is valid, or to~0 otherwise
(cf., \Cref{tbl:esempio}).

The iterative constraint propagation algorithm 
uses few additional data structures:
\begin{itemize}
    \item $\mathit{\sval}$: A list of the indexes of variables with domain reduced by the last iteration.
    \item $\mathit{\ssup}$:  A list of the indexes of the variables  
    such that $|\dom(x)| > 1$.
    \item $\Delta$: for each $x \in \var(c)$,  $\Delta_{x}$ is the set of values removed from $\dom(x)$ in the last iteration.
    \item $\mathit{lastSizes}$: storing the current sizes of the domains: $\mathit{lastSizes} = [\,|\dom(x_1)|,\dots,|\dom(x_n)|\,]$.
\end{itemize}

For a given table constraint $c$, Algorithm~\ref{alg:enf} enforces GAC first by initializing the data structures described earlier (lines~1--3), and then by calling 
Algorithms~\ref{alg:updateTable} and~\ref{alg:filter}. 
Assuming that for each variable $x_i$ the values in $\Delta_{x_i}$
have been removed from $\dom(x_i)$, Algorithm~\ref{alg:updateTable} updates $\mathit{currTable}$ by collecting the bits corresponding to all (still) valid tuples.
To minimize the number of operations, a choice is made between a \emph{reset-based update} and an \emph{incremental update}.
The number of bitset operations required by the two options depends on the size of~$\Delta_{x_i}$   (see the paper by \citeN{main} for a detailed description).
The choice is made in line~3:
If $|\Delta_{x_i}|$ is smaller than the current domain size of $x_i$, $mask$ is used (lines~4--6) to collect bits corresponding to tuples that have become unsupported (because of the removed elements in $\Delta_{x_i}$). 
Alternatively, in lines~8--9, the bits that will compose the updated $\mathit{currTable}$ 
are gathered by inspecting $\dom(x_i)$.
After the update of $\mathit{currTable}$  Algorithm~\ref{alg:filter} 
 filters the current domains using the outcome of Algorithm~\ref{alg:updateTable}. 
 
Enhancements, not discussed here, such as the usage of residues \cite{main}, can be applied  to improve efficiency of the filtering step.

{\small
\begin{algorithm}[tb]
\small
    \SetKwInOut{Input}{input}
    \SetKwInOut{Output}{output}
    \Input{{\it supports}, {\it currTable}, $\dom$, {\it lastSize}}
    \For(\tcp*[f]{For all variables with non-singleton domain}){$i \in \ssup$}{
        \For(\tcp*[f]{Check the support for $a$ in a tuple of the table}){$a \in \dom(x_i)$}
        {
         \If(\tcp*[f]{If not supported remove the value}){$currTable \;\& \;\mathit{supports}[x_i,a] = \vec 0$}{
           $\dom(x_i) \leftarrow \dom(x_i) \setminus \{a\}$        
            }
        }
        $\mathit{lastSize}[i] \leftarrow |\dom(x_i)|$ \tcp*[r]{Update the sizes}
    }
    \caption{filterDomains()}
    \label{alg:filter}
\end{algorithm}
}

\subsection{The {\sc MiniCP} solver and GPU}

Since the late Eighties several efficient  constraint solvers have been developed for Constraint (Logic) Programming. For the scope of this paper we have decided to use {\sc MiniCP}~\cite{minicp}, a
relatively new and open-source  solver devised to ease development of extensions. In particular, our extensions build on {\sc MiniCPP} \cite{minicpp}, a C++ implementation of {\sc MiniCP} that can easily include CUDA C++ code (\citeNP{cumulative};\citeyearNP{DBLP:journals/logcom/TardivoDFMP23,DBLP:conf/cp/TardivoMP24}).
New global constraints can be integrated in {\sc MiniCP} by creating a class under {\small\url{/fzn-minicpp/global\_constraints/}} which extends the \texttt{Constraint} class. Such a class, which will serve as the propagators for the new global constraints, must include, alongside a constructor, a procedure, \textit{propagate()}. 
The \textit{propagate()} method implements the propagation algorithm for the new constraint. 
To provide a simple mechanism to use the GPU-accelerated propagators, we rely on the MiniZinc annotation introduced by \citeN{DBLP:journals/logcom/TardivoDFMP23}.
Specifically, a table constraint annotated with \texttt{::gpu} 
(i.e., a constraint declared as: \texttt{constraint table(...){::}gpu;\,} in the input model) is propagated using the GPU-accelerated implementation in place of the standard one.

\section{Serial implementation of CT propagation}
\label{sec:serial}\label{section:serialImplementation}

We briefly outline a serial implementation of the Compact-Table propagator described in Section~\ref{sub:CT}, conceived to be integrated in  {\sc MiniCP} as a new global constraint. 
We will refer to this implementation simply as CT.
As mentioned, while enforcing GAC, the (serial) Algorithm~\ref{alg:updateTable} gathers the set of valid tuples by performing either a \emph{reset-based} update or an  \emph{incremental update}. 
{\sc MiniCP} provides no built-in functions to keep track of the domain changes, so, to avoid the computation of~$\Delta$ we opted to use only the reset-based strategy, in all our implementations. 
Indeed, the different update strategies were introduced to minimize the number of computations needed to update the $\_currentTable$, at the cost of calculating and maintaining information about the previous domains. Our objective is to implement a GPU propagator, calculating and copying $\Delta$ from host to device can be less effective than using a simpler strategy, offloading the additional computation steps to the device.

\medskip
\noindent\textsc{Data structures}: 
A specific class $Table$ stores  the table instance at hand and defines the methods for enforcing GAC.
{\sc MiniCP} itself manages all the data related to the variables.
The main data structures used closely match those mentioned in Section~\ref{sub:CT}: 
\begin{itemize}
    \item $\mathit{\_currTable}$: A bitset implementing the $\mathit{currTable}$ described in Section~\ref{sec:bg}.
    
    \item $\mathit{\_currTableSize}$: An integer storing the number of words of $\mathit{\_currTable}$.
    \item $\mathit{\_supports}$: The compact support table, namely a Boolean matrix stored as an array of 32-bit words. Each word of $\mathit{\_supports}$ compactly stores 32 Boolean values. 

    \item $\mathit{\_supportSize}$: An integer storing the number of rows of the support matrix, $\mathit{\_supportSize}= \sum_{x \in \var(c)} |\dom(x)|$.
    \item $\mathit{\_s\_val}$, $\mathit{\_s\_sup}$: Lists of integers (variable indexes), corresponding to $\sval$ and $\ssup$.
    \item $\mathit{\_supportJmp}$: An array of integers. 
    For each variable $x_i$, $\mathit{\_supportJmp[i]}$ is the index of the first cell in $\mathit{\_supports}$ where a value for $x_i$ occurs. 
    For example, for the constraint in Table~\ref{tbl:esempio} 
    we have $\mathit{\_supportJmp[0]=0}$, $\mathit{\_supportJmp[1]=3}$, and $\mathit{\_supportJmp[2]=6}$. Note that 
    $\mathit{\_supportJmp}$  is a constant vector and it is used to speed up the accesses to $\mathit{\_supports}$. 
    \item $\mathit{\_variablesOffsets} $:  A vector of integers of length $n$. It contains the first value in the initial domain of each variable. It is used in accessing the rows of $\mathit{\_supports}$. As an example let $x_i$ be a variable with a domain $dom(x_i)=[90,120]$, then $\mathit{\_variablesOffsets[i]=90}$. 
\end{itemize}

Let us remark that the data structure $\mathit{lastSizes}$ used in Algorithm~\ref{alg:enf} is not used since we rely on the {\sc MiniCP} built-in method $\mathit{changed}()$
to populate $\_s\_val$.
This method returns \emph{true} when invoked on a variable whose domain changed during the previous iteration.
Moreover, since  we adopt the reset-based update approach, we do not  use the structure~$\Delta$.

\medskip
\noindent\textsc{Procedures}: 
It suffices specializing the class \texttt{Constraint} to extend {\sc MiniCP}. 
We implemented a new class \emph{Table} whose constructor allocates and initializes all the data structures described earlier. 
It also checks if each variable in the constraint is supported by at least one tuple, otherwise unsatisfiability is reported.
The method $\mathit{propagate}()$  corresponds to three methods $\mathit{enforceGAC}()$, $\mathit{updateTable}()$, and $\mathit{filterDomains}()$ that, except for the indexed access to  $\mathit{supports}$, based on $\mathit{\_supportJmp}$, are one-to-one implementations of Algorithms~\ref{alg:enf}--\ref{alg:filter}.

\section{GPU accelerated CT propagators}
\label{sec:parallel}
The main engine of {\sc MiniCP} performs the Constraint-Based Search in the usual manner, essentially by alternating search and propagation phases, possibly involving backtracking steps. 
Intuitively, whenever the GAC enforcing procedure is triggered, the corresponding method is responsible to start the propagation and report the results to the main engine.
The possibility of extending {\sc MiniCP}  with a new propagator is transparent to the way it is implemented.  
Indeed, provided that the propagator adheres to the interface with the solver, there is no requirement on how its computation is performed.
So, the same mechanism used to add a serial propagator can be used to plug in a GPU-empowered method. Clearly, this method is also responsible to transparently offload the computation on the GPU and to suitably manage the related data transfers.

Recall that each computation on the device is described as a collection of concurrent threads, each executing the same function. Such a function is called \emph{kernel}, in CUDA terminology. Threads are hierarchically organized in equally sized blocks that are in turn structured in a grid (see Figures~\ref{fig:updateTableGPU},~\ref{fig:reduce}, and~\ref{fig:filterDomainsGPU}). 

We implemented parallel counterparts of $\mathit{updateTable}()$ 
and $\mathit{filterDomains}()$, composing them into three variants of the propagator:
\begin{itemize}
\item $CT_{CU}^{u}$: only $\mathit{updateTable}()$ is computed on the device. 
\item $CT_{CU}^{f}$: only $\mathit{filterDomains}()$ is computed on the device. 
\item $CT_{CU}^{uf}$: both update and filtering are computed on the device. 
\end{itemize}

All implementations extend the class $Table$ described earlier.
The constructors of the derived classes allocate and initialize the data structures on the GPU. 
When propagation is triggered, relevant data is transferred to the device, the computation is run, and the results are retrieved.

\subsection{The propagator \texorpdfstring{\mbox{$CT_{CU}^{u}$}}{CTCUu}}

\begin{figure}[b]
    \includegraphics[width=0.90\linewidth]{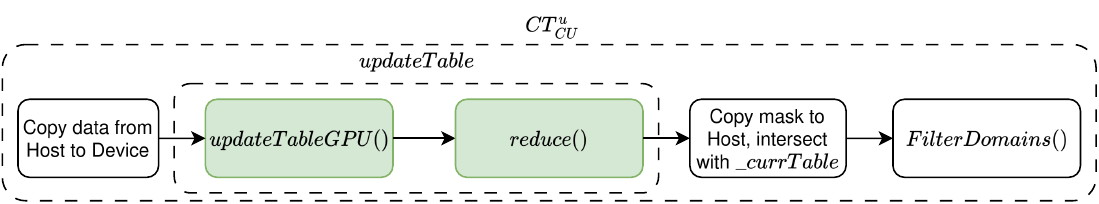}
    \caption{Overview of the execution flow of $CT_{CU}^{u}$. Kernels are depicted by green activities.}
   \label{fig:ctu}
\end{figure}

In this propagator the execution flow of $\mathit{enforceGAC}()$, closely matches its serial counterpart. 
First, the contents of $\mathit{\_currTable}$ and $\mathit{\_s\_val}$ are copied to the device, together with the variables' domains.
The kernels responsible for updating the current table, i.e., $\mathit{updateTableGPU}()$ and $\mathit{reduce}()$, are subsequently launched, see Figure \ref{fig:ctu}. The outcome of this step is the same mask computed by the original $\mathit{updateTable}()$ procedure, but it is written in GPU's global memory. It is then copied back to the host and combined bitwise with $\mathit{\_currTable}$. Finally, the filtering procedure on the host is invoked.

The propagator $CT_{CU}^{u}$ uses some auxiliary structures that do not have a counterpart on the serial side.
They are:
\begin{itemize}
     \item $\mathit{\_supportsT\_dev}$: An array of integers storing the transposed support matrix. It is used by the kernel $\mathit{updateTableGPU}()$ for updating the current table. 
     The transpose is exploited to make threads of the same warp access adjacent memory words.
    \item $\mathit{\_vars\_dev}$: An array of $ \lceil\mathit{\_supportSize/32}\rceil$ integers. It stores the concatenation of the bitmaps representing the different variable domains. It is the counterpart of an host side variable ($\mathit{\_vars\_host}$) storing variable domains.  The content of $\mathit{\_vars\_host}$ is copied to the device at each propagation.   

    \item $\mathit{\_tmpMasks}$: An array of integers of size $\mathit{\_currTableSize\times n}$. It is a linearized representation of the support matrix, in which each row (of size $\mathit{\_currTableSize}$) is associated to a variable.
    Namely, for each variable $x_i \in \_s\_val$,  the row $\mathit{\_tmpMasks[i]}$ stores the  $\mathit{bitwise}${-}$\mathit{OR}$ of the rows $\mathit{\_supports}[x_i,a]$ for all $a \in dom(x_i)$. Such a structure  stores the result of the kernel $\mathit{updateTableGPU}()$ and is subsequently processed by a $\mathit{reduce}()$ kernel  (see below).
\end{itemize}

\begin{figure}[tb]
 \centerline{\includegraphics[width=1.05\linewidth]{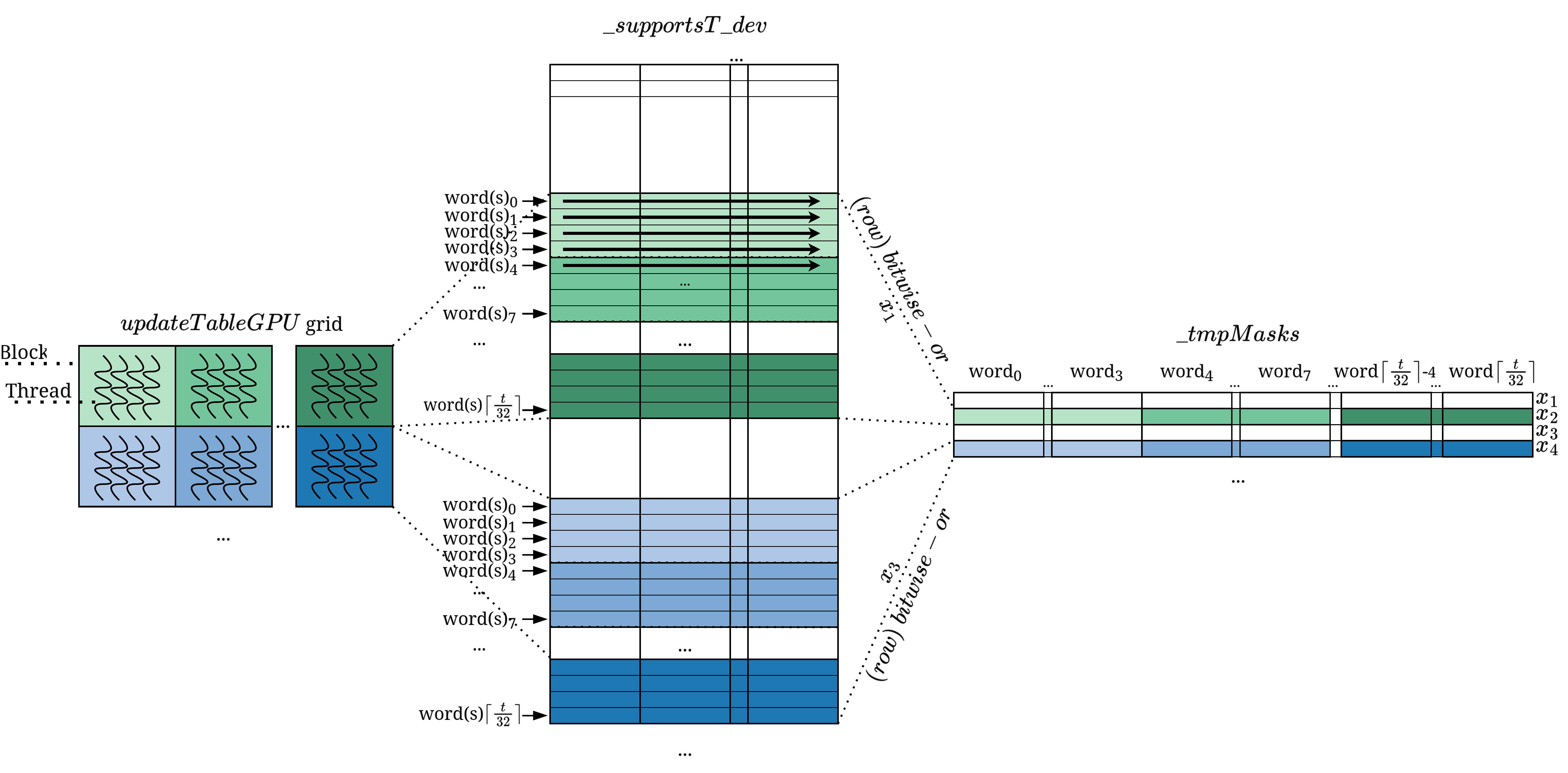}}
 \caption{Overview of the parallel reduction done on $\_supportsT\_dev$, where $\_s\_val=\{x_1,x_3,\ldots\}$. Different colors indicate which words are accessed by the threads in the block. Arrows depict the direction of the parallel reductions and $\lceil \frac{t}{32} \rceil$ is a shorthand for $\_currTableSize$ (see Section~\ref{sec:bg}).}
  \label{fig:updateTableGPU}
\end{figure}

\begin{figure}[tb]
 \centerline{\includegraphics[width=1.05\linewidth]{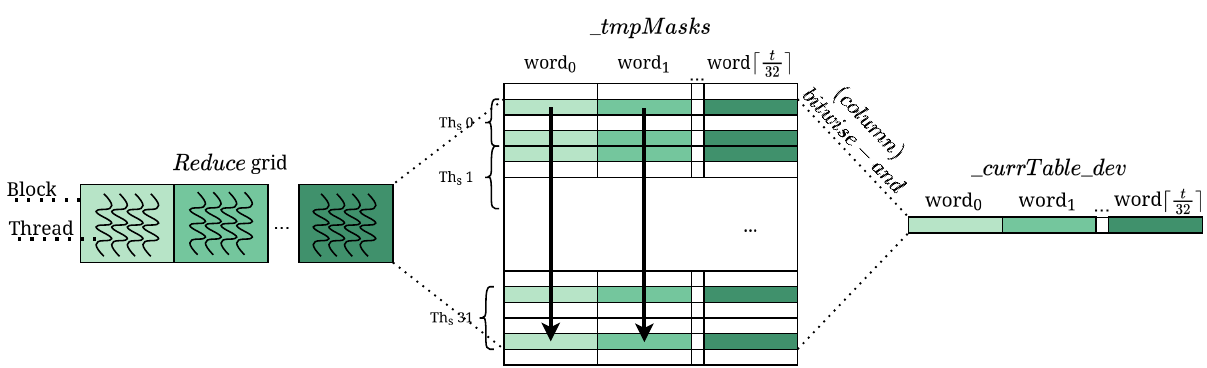}}
    \caption{Overview of the parallel reduction done on $\mathit{\_tmpMasks}$ by 
    $\mathit{reduce}()$. Different colors of the blocks indicate which words are accessed by the threads in the block. Arrows show the direction of the parallel reductions while $\lceil \frac{t}{32} \rceil$ is a shorthand for $\_currTableSize$ (see Section~\ref{sec:bg}).}
   \label{fig:reduce}
\end{figure}

As concerns the kernel $\mathit{updateTableGPU}()$, the grid launched has $\lceil \_currTableSize/4 \rceil \times |\_s\_val|$ blocks.
That is, a block of 128 threads for every four (32-bit) words used by $\mathit{\_currTable}$ and for each changed variable is launched. Note that the division of work between blocks/threads (and thus between streaming multiprocessors of the GPU) is dynamic and scales with the size of the problem (i.e. the number of rows in the table as well as the scope size).

Each row of the grid is related to a variable reduced during the last iteration, i.e. the $i$-th row of the grid is related to the $i$-th variable in $\_s\_val$, assume it to be $x_j$. 
Each block in the row is responsible for updating four (32-bit) words of the table mask accessing the \emph{transposed} supports related to $x_j$. 
Each block of 128 threads can be seen as a small sub-grid of $32 \times 4$ threads, where each of the 32 rows considers a (different) slice of $dom(x_j)$. Each row of this sub-grid computes, via a $\mathit{bitwise}${-}$\mathit{OR}$, the partial mask that needs to be added to $\mathit{\_currTable}$. 
Such a mask, computed in shared memory is then copied in global memory in the $i$-th row of $\mathit{\_tmpMasks}$ (see Fig.~\ref{fig:updateTableGPU}).

Once $\mathit{updateTableGPU}()$ is terminated, the kernel $\mathit{reduce}()$ is launched. 
It considers all and only the rows of $\_tmpMasks$ related to changed variables (i.e. those in $\_s\_val$) and performs a $\mathit{bitwise}${-}$\mathit{AND}$ operation along the columns.
To this aim, a grid of $\mathit{\_currTableSize}$ blocks of 32 threads each is launched. Each block processes a different column of $\mathit{\_tmpMasks}$.
Similarly to $updateTableGPU()$, also for the $\mathit{reduce}()$ kernel the division of work is dynamic and scales with the number of rows in the table.

Consider the $i$-th block of the $reduce$ grid, each thread of such a block looks at a word (same column) in different rows of $\mathit{\_tmpMasks}$ performing a $\mathit{bitwise}${-}$\mathit{AND}$ operation.
Once all the rows have been considered, all the threads of the same block perform a parallel reduction (applying a $\mathit{bitwise}${-}$\mathit{AND}$) to produce a single 32-bit word per block (see Fig.~\ref{fig:reduce}).
After this kernel is executed, the obtained mask is copied back to the host.

\subsection{The propagator \texorpdfstring{\mbox{$CT_{CU}^{f}$}}{CTCUf}}
\begin{figure}[tb]
    \includegraphics[width=0.90\linewidth]{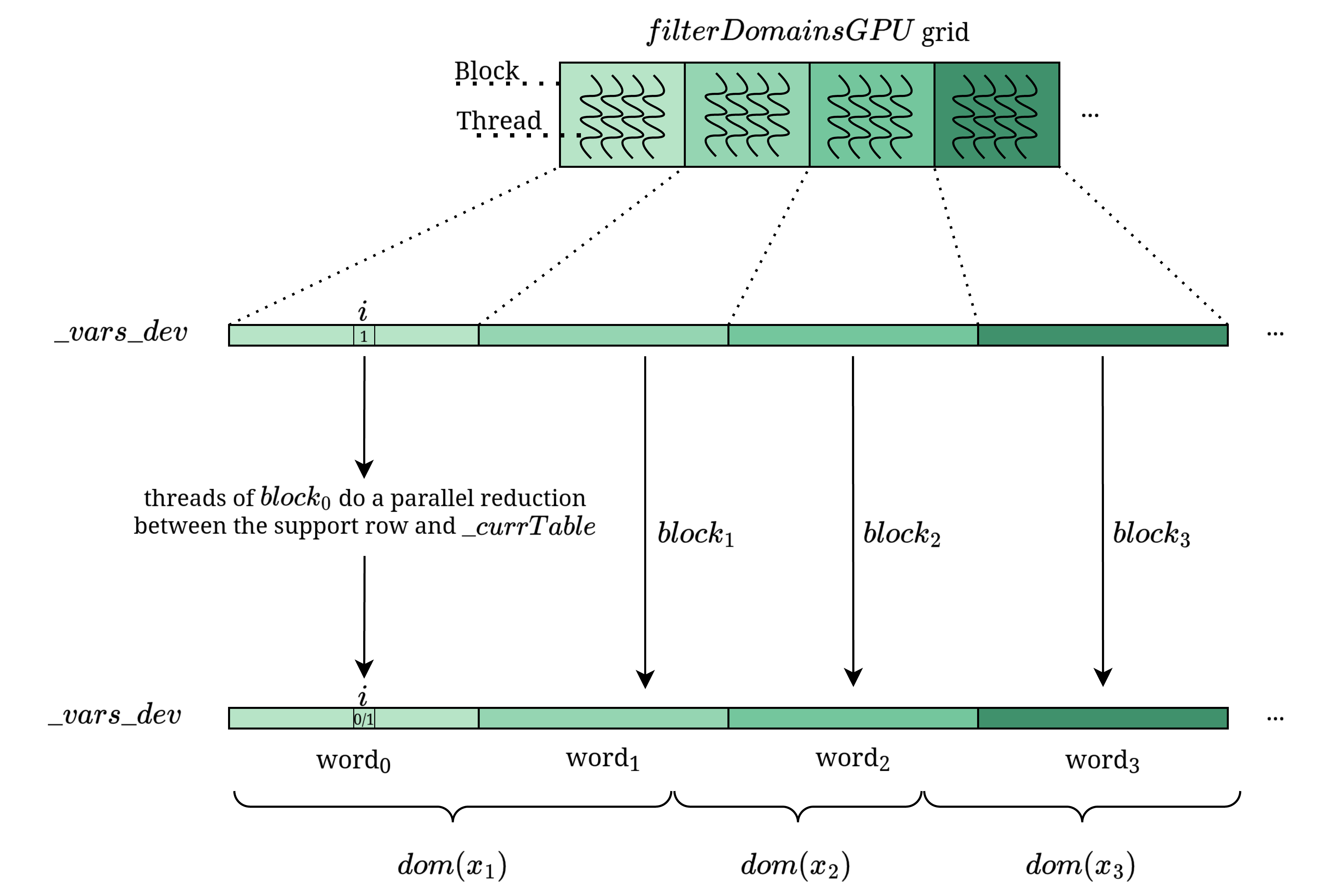}
    \caption{Overview of the data each block considers in the kernel $\mathit{filterDomainsGPU}()$.}
    \label{fig:filterDomainsGPU}
\end{figure}

In the case of the propagator $CT_{CU}^{f}$, the update of $\mathit{\_currTable}$ is carried out by the host while the filtering procedure is carried out on the device by means of the $\mathit{filterDomainsGPU}()$ kernel.
Such a kernel (see Fig.~\ref{fig:filterDomainsGPU}) is responsible for providing the host, for each variable in the scope, the values to remove from their respective domains.
The initialization and data structures used are the same as for $CT^u_{CU}$, with the exception of a new array $\mathit{\_vars\_to\_remove\_host}$ on the host side. Such an array, of size $\lceil \mathit{\_supportSize}/32\rceil$, stores  the result of the $\mathit{filterDomainsGPU}()$ kernel.

As concerns the  $\mathit{filterDomainsGPU}()$ kernel, the number of blocks launched is $\lceil\mathit{\_supportSize}/32\rceil$, each consisting of 32 threads. 
The division of work is thus dynamic and depends on the domain size of the variables in the scope.
Each block is responsible for processing a single 32-bit word of the variable domains stored in $\mathit{\_vars\_dev}$.
All threads in the same block check whether the same value is still present in the domain of the variable $x_j$ considered. If such is the case, a parallel reduction on the 32 threads is launched on the respective support row, computing a $\mathit{bitwise}${-}$\mathit{AND}$ with the current table. 
After the parallel reduction, the $i$-th bit of a shared word in the block is set to~1 if the value considered must be removed from the variable, or to~0 otherwise.
The kernel thus computes an array of size $\lceil\mathit{\_supportSize}/32\rceil$, in which the bit in position $\mathit{\_supportJmp[j]}+i-\mathit{\_variablesOffsets[j]}$ is set to~$1$ if and only if the $i$-th value of the variable~$x_j$ must be removed from its domain. Each block is responsible for writing one word in $\_vars\_dev$.  
Such an array is then copied back to the host into $\mathit{\_vars\_to\_remove\_host}$ and the values are removed from the domains of the respective variables.

As an example consider a table with 1000 rows and suppose the scope has two variables, $x_1$ and $x_2$, respectively with $dom(x_1)=\{100,\ldots,196\}$ and $dom(x_2)=\{1,\ldots,100\}$. A total of 7 blocks, of 32 threads each, are launched. Consider now the second block (index~1), all 32 threads of such a block  check the 32 bits responsible for keeping track of whether, for the variable $x_1$, the values in the set $\{132,\ldots, 163\}= \{\mathit{\_variableOffsets[0]}+32,\ldots,\mathit{\_variableOffsets[0]}+63\}$ are still present in $dom(x_1)$.

\subsection{The propagator \texorpdfstring{\mbox{$CT_{CU}^{uf}$}}{CTCUuf}}
This last implementation combines the kernels designed for  $CT_{CU}^{u}$ and $CT_{CU}^{f}$. 

The kernel $\mathit{filterDomainsGPU}()$ is called after
$\mathit{updateTableGPU}()$ and $\mathit{reduce}()$, so that both the table update and the domain filtering are carried out by the device.
The data structures used are those introduced in both $CT^u_{CU}$ and $CT^f_{CU}$.

\section{Experiments}\label{sec:experiments}

We report the computational behaviour of 
the serial $CT$ and the two
parallel propagators $CT_{CU}^{f}$ and 
$CT_{CU}^{uf}$ and on a comparison with Gecode, the default solver of 
Minizinc.

$CT^{u}_{CU}$ results are omitted both because this implementation sometimes underperforms relative to the serial counterpart, and to avoid shifting the focus away from the main results, and because of limited space.
One reason why $CT^{u}_{CU}$ on some instances does not keep up with the serial implementation is that the host function $\mathit{updateTable}()$ is very efficient and the two kernels $\mathit{updateTableGPU}()$ and $\mathit{reduce}()$, while being faster than it, cannot overcome alone the overhead introduced by freqently moving data and control to/from the GPU.

The choice of using MiniZinc and not use CLP(FD) was driven by the availability of a compiler from MiniZinc to FlatZinc and {\sc MiniCP} supporting FlatZinc. For this reason we encoded our tests in MiniZinc. Such a language is widely adopted in the constraint programming community and used in the MiniZinc Challenges \cite{mznChallenge}.
For completeness, we recall that also compilers from CLP languages to FlatZinc have been developed \cite{clpfd}.

\subsection{The benchmark suite}
\label{sub:testGen}

We report here on the experimentation we conducted using three collections of instances.
The  instances of the first set involve a single table constraint and a single linear equation on \emph{some} of the variables in the scope of the table constraint, as in the following example: 
\lstset{basicstyle=\ttfamily\small}
\begin{lstlisting}[label={lst:lin}]
  var 1..10 : x1; var 1..10 : x2; var 1..10 : x3;  var 1..10 : x4;
  var 10..20 : x5;
  array [int,int] of 1..10: t = [|1,1,1,1 | 1,2,1,2| 2,3,2,3|...|];
  constraint table([x1,x2,x3,x4], t) :: gpu;
  constraint 2*x1+3*x2+4*x3=120;
  solve satisfy;
\end{lstlisting}

  Note that we admit variables in the linear constraint not occurring in the table constraint. 
We generated two batches of instances, called $LIN\_B$ and $LIN\_EB$ resp., differing in the number of variables and tuples (see Table~\ref{tbl:instances}).
  The (synthetic) instances of this benchmark are based on a simple encoding of the bounded knapsack problem.
  More specifically, in the bounded Knapsack problem there are $k$ items and a number representing the capacity of the knapsack. Each item is associated with a weight, a profit, and a maximum number of times the object can be included in the knapsack.  The goal is to select the number of instances of each item maximizing the profit which can fit the knapsack \cite{knapsack}. 
  This problem can be modeled using two constant arrays ($\textit{profit}$ and $\textit{weight}$) and a variable array representing, for each object, the number of times it is put in the knapsack. A table describing all possible knapsack configurations (i.e., the variables in the scope are the elements of the variable array) is defined. Additionally, a constraint requiring the weights of the objects considered to fit the knapsack is added (this is the linear equation mentioned earlier).

The tables in the $LIN\_B$ and $LIN\_EB$ tests may not involve all variables declared in the instance and do not fix a specific solving strategy, so such tests may trigger  different numbers of propagations when solved by different solvers (e,g., Gecode and $CT^{uf}_{CU}$).
  In order to have a fair comparison between our implementations and Gecode,
  we introduced a second set of instances where both the linear equation and the table constraint involve all variables occurring in the instance.
  Moreover, in each solver, the solving strategy has been fixed as follows:

\lstset{basicstyle=\ttfamily\small}
\begin{lstlisting}
solve :: int_search(
  [x0, x1, x2, ..., x_n],   % The variables in the scope
  input_order,              % Variable selection strategy
  indomain_max,             % Value selection strategy
  complete                  % Search completeness
) satisfy;
\end{lstlisting}

\begin{table}[tb]
    \centering
    \begin{tabular*}{0.6\textwidth}{@{\extracolsep{\fill}}cccc}
        Benchmark & Variables & {Max Domain Size} & Tuples \\ 
        \midrule
        $LIN\_B$ & 80-150 & 600 & 5000-10000\\ 
        $LIN\_EB$ & 100-200 & 800 & 5000-15000
    \end{tabular*}
    \caption{Overview of table constraint features in the first test set.}
    \label{tbl:instances}
\end{table}

Finally, we experimented with a set of instances of the $\textit{Orienteering problem (OP)}$ or \textit{Single-Agent Path Planning with Reward Maximization}~\cite{orientiring}.
The OP seeks a path for an agent starting and ending at specified locations, imposing a bound $D$ on the length of the path and maximizing the reward accumulated by visiting the various sites along the path, up to a threshold~$C$.
The problem can be formalized as a problem on an undirected graph $G = \langle V, E\rangle$, with nodes $V=\{v_1,\ldots,v_k\}$ and edges $E$, such that, for each edge $(v_i,v_j)$,  $w_{i,j}$ denotes its length and for each node $v_i$,   $\pi_i$ is a profit value associated to~$v_i$. 
Given such $G = \langle V, E\rangle$, a starting and an ending nodes $v_s, v_e \in V$, 
a length bound $D \in \mathbb{R}$, a set $\Pi=\{\pi_1,\ldots,\pi_k\}$ of profit values for nodes, and a profit bound $C \in \mathbb{R}$, an instance of OP asks to determine a path $P=\langle (v_{p_1},v_{p_2}),(v_{p_2},v_{p_3}),\ldots,(v_{p_k-1},v_{p_k})\rangle$ (with $v_{p_1}=v_s$ and $v_{p_k}=v_e)$ such that $\sum_{(v_{p_i},v_{p_j}) \in P}w_{{p_i},{p_j}}\le D$.
Such a path must also maximize the  profit collected by visiting the nodes, subject to the bound $C$, namely, $(\sum_{(v_{p_i},v_{p_j})\in P} \pi_{p_i}) + \pi_{p_k} \leq C$.

We used a collection of instances of OP in which $G$ is a 4-connected grid (a graph in which each node is connected to at most four other nodes) with $w_{ij} = 1$ for each $(v_i,v_j) \in E$. 
Hence, a path can exit a node in one among (at most) four possible directions.
We did two additional changes, w.r.t.\ the original definition of~OP:
\begin{itemize}
\item Each node  can either contain a non-negative profit or be an obstacle, in which case it is not traversable by any path.
\item The profit corresponding to any node in the path of the agent is collected once. 
\end{itemize}

To model this problem, we use $\lceil \frac{D}{2} \rceil$ table constraints, each describing agent position and accumulated reward across two consecutive time-steps. If $G$ is a grid of size $r\times s$, then each table contains $2\times r \times s$ columns. In each row, only two cells 
are assigned values between $1$ and $D$, representing agent position and collected reward.
Instances with grid size from $7\times7$ to $10\times10$ have been generated (see Fig.~\ref{fig:orienteering}).

\subsection{Results}
\label{sub:tests}
We now report the results considering the \emph{solve time} statistic provided by MiniZinc on the instances described in Section~\ref{sub:testGen}.
Such tests have been executed on both CT\mbox{{}$_{CU}^{f}$} and CT\mbox{{}$_{CU}^{uf}$} implementations on a system equipped with an Intel Core i7-13700KF at base clock $3.4$ GHz (up to $5.4$ GHz in boost), 128 GB of DDR4 RAM and a NVIDIA GeForce
RTX 4090 with 16K CUDA cores, grouped into 128 streaming multiprocessors, running at $2.23$ GHz (up to $2.52$ GHz). On the software side, the system runs Ubuntu 24.04 and CUDA 12.7.
All tests have been executed with 15 minutes timeout.
Barplots summarizing the serial ($CT$) and CUDA solve times are now presented. 
The same results in tabular form are available in the appendix. 
Such tables show also the number of propagation performed for each test, allowing to calculate additional metrics, such as the performance gain in terms of propagations per second.
Propagation times of some tests
are analyzed more in detail in Section~\ref{subsub:propagations_detail}.

\begin{figure}[tb]
    \includegraphics[width=0.90\linewidth]{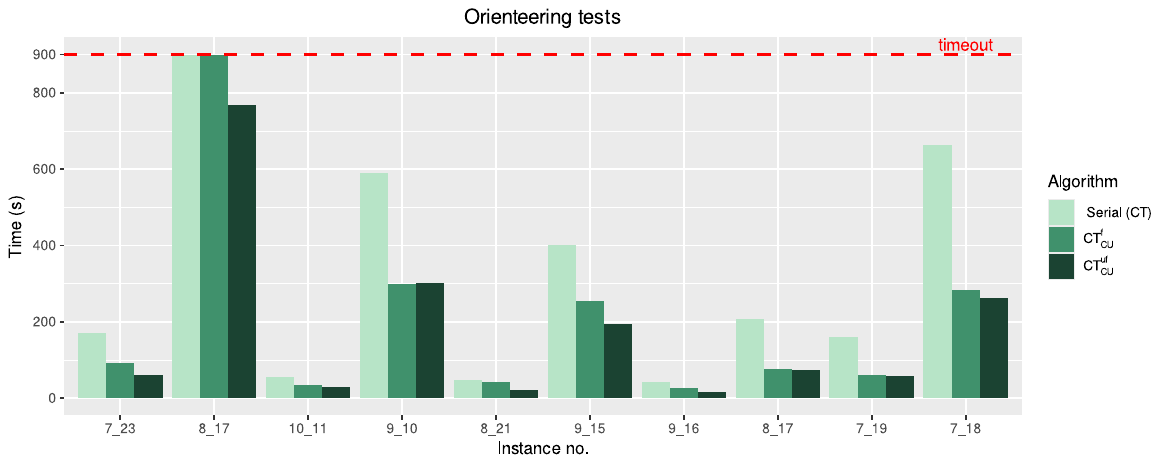}
    \caption{Barplot comparing serial and CUDA solve times for the \textit{OR} instances. The grids considered are squared, the instance named $r\_C$ is an $OR$ instance with a $r\times r$ grid and profit bound $C$.}
    \label{fig:orienteering}
\end{figure}

\begin{figure}[tb]
    \includegraphics[width=0.9\linewidth]{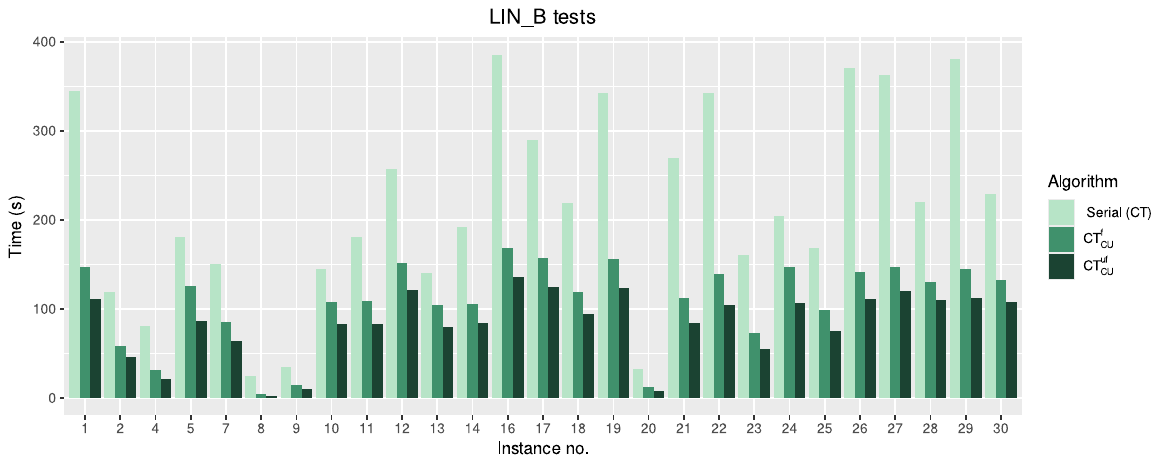}
    \caption{Barplot comparing serial and CUDA solve times for the first batch of instances. Test instances where all implementations timed out are omitted.}
    \label{fig:sumUp4090_b}
\end{figure}

\begin{figure}[tb]
    \includegraphics[width=0.9\linewidth]{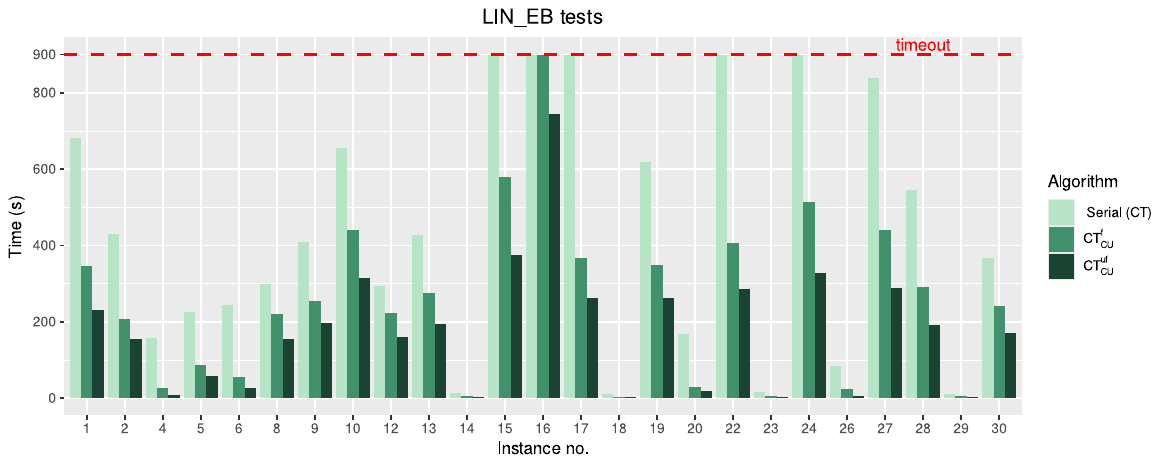}
    \caption{Barplot comparing serial and CUDA solve times for the second batch of instances. Test instances where all implementations timed out are omitted.}
    \label{fig:sumUp4090_eb}
\end{figure}

From Figures~\ref{fig:sumUp4090_b} and~\ref{fig:sumUp4090_eb} it can be observed that the kernels responsible for the update step scale well, taking advantage of the computational power of the GPU. 
Indeed, for all sufficiently large instances, $CT^{uf}_{CU}$ outperformed $CT^{f}_{CU}$, achieving an average speedup of $2.88$ compared to $2.11$ on $LIN\_B$ instances, and $4.35$ compared to $2.52$ on $LIB\_EB$.
 Figure \ref{fig:orienteering} shows an overview of the solving times and speedups for the \emph{Orienteering} test set. Notice that while such models involve several other constraints in addition to multiple table instances, the speedups remain consistent with those of \emph{LIN\_B} and \emph{LIN\_EB} instances.


Kernel execution profiling showed that during the tests performed, the GPU capabilities were not fully exploited.
In all the tests, the average GPU utilization was roughly $45\%$.
This is because the amount of work offloaded to the GPU is often not enough to saturate all streaming multiprocessors. (To saturate the capabilities of the RTX 4090, much larger instances would be needed, involving much larger tables.)
This supports the fact that the good performance achieved using the RTX 4090  is also expected when using GPUs equipped with less streaming multiprocessors.

\enr{Note that forcing a finer division of work for the sole purpose of using all available streaming multiprocessors often results in performance degradation. In such cases, in fact, required work synchronization and additional data exchange generate overhead that outweighs the benefits of parallel computation.}

\enr{As concerns memory transfers, the time taken by memory copies from the host to the device and vice versa is significant when compared to the pure processing time.
In part, this is due to the fact that the $CT$ algorithm is very efficient. On average, on $LIN\_B$ and $LIN\_EB$ instances, the copy of data to the device can take up to $50\%$ of the kernel time, while the copy of the results back to the host can take up to $80\%$ of the kernel execution time.}    
\subsubsection{\texorpdfstring{$CT^{uf}_{CU}$ propagations in detail}{CT-CU-uf-propagations in detail}}
\label{subsub:propagations_detail}
\begin{figure}[tb]
\centerline{\includegraphics[width=1\linewidth]{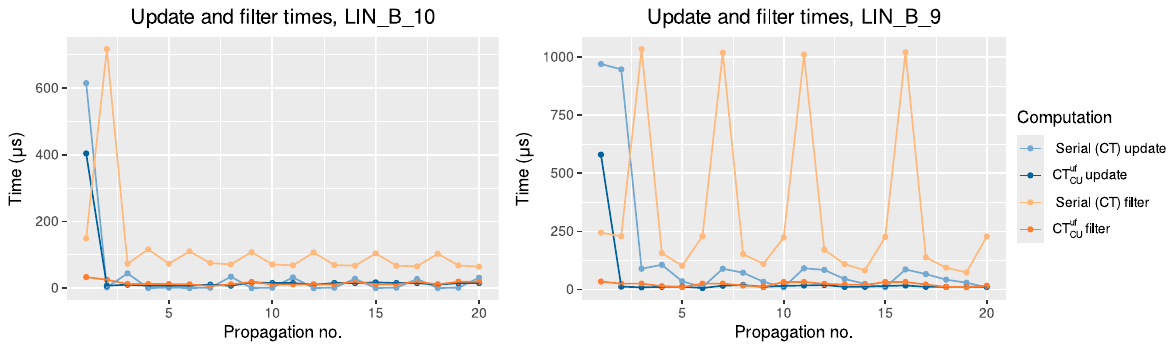}}
    \caption{Series plots of the first 20 $CT$ and $CT^{uf}_{CU}$ propagation times for $\mathit{TEST\_B\_10}$ and $\mathit{TEST\_B\_9}$. The times are grouped for the update and filter procedures.
    }
\label{fig:test_9_10_series}
\end{figure}

In this section we briefly analyze the behavior of $CT^{uf}_{CU}$ executed on the first set of instances. 
We do so by analyzing the first 20 propagations of $LIN\_B\_9$ and $LIN\_B\_10$, which represent the two most representative behaviors observed in experimenting with the $LIN\_B$ and the $LIN\_EB$ instances. 
In the first scenario, $\ssup$ has a stable size, making the serial filtering times quite constant.
In contrast, the second case is characterized by high variability in the size of $\ssup$, leading to spikes in the serial filtering times, see Fig.~\ref{fig:test_9_10_series}. 
In both scenarios however, the times recorded for the filtering kernel remain almost constant, this is due to the fact that the size of the grid launched is independent from $|\ssup|$.
As it can be seen from Fig.~\ref{fig:test_9_10_series}, concerning the first kind of instances, the filtering process is the procedure leading to the most sensible improvements when parallelized. 

For what concerns the procedure responsible for updating $currTable$, Fig.~\ref{fig:test_9_10_series} compares the sum of the times of kernels $\mathit{udpateTableGPU}()$ and $\mathit{reduce}()$ (in $CUDA\; update$) with those of $\mathit{updateTable}()$.
Such times, when paired with a small number of propagations, do not lead to a significative difference.
Moreover, the sum of $\mathit{udpateTableGPU}()$ and $\mathit{reduce}()$ times, due to the grids not presenting a constant size, present the same time trends as the serial counterpart.

To have a direct comparison between the filtering and update steps performed on GPU and their serial counterparts, the times considered in Fig.~\ref{fig:test_9_10_series} do not include the copy and retrieval of the data nor the time taken to dump the domains in $\mathit{\_vars\_host}$.

\begin{figure}[tb]
    \includegraphics[width=0.90\linewidth]{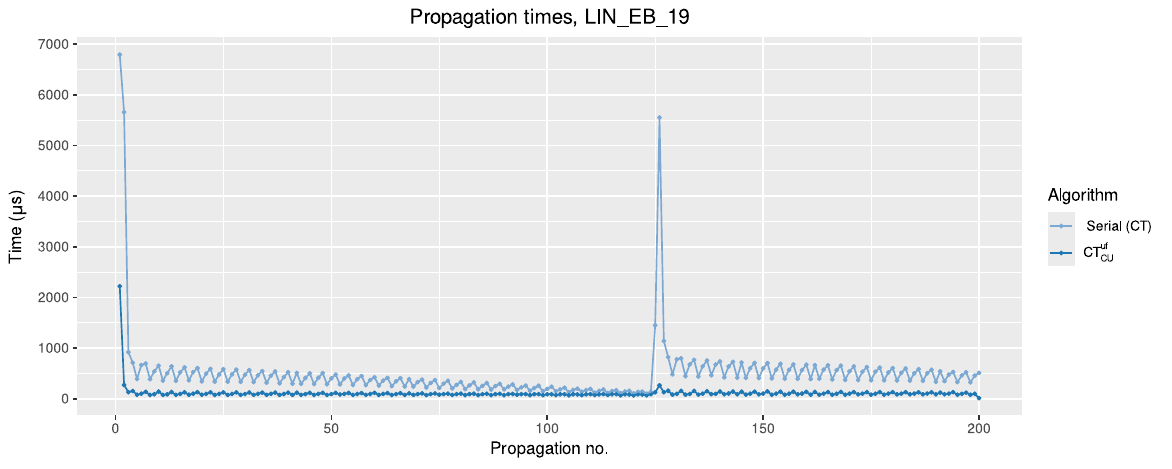}
    \caption{Series plot of the first 200 $CT$ and $CT^{uf}_{CU}$ propagation times for $\mathit{TEST\_EB\_19}$.}
    \label{fig:test_26_series}
\end{figure}

Finally, we discuss the times spent by the overall propagations in $CT$ and $CT^{uf}_{CU}$. For what concerns the serial propagations the time only includes the update of $\_currTable$ and filtering. On the other hand, for $CT^{uf}_{CU}$ the time accounts for the kernels, the dump of the domains into $\_vars\_host$, the copy and retrieval of the data. Fig.~\ref{fig:test_26_series} shows the timings of the first 200 propagations of $LIN\_EB\_19$. Such an instance has been chosen since already in early propagations it presents a significant backtracking phase. As can be seen, once the valid tuples and domains start to reduce, the serial propagation times become very close to the $CT^{uf}_{CU}$ ones. In some instances requiring many propagations toward the bottom part of the search tree, it can also happen that the serial propagations run faster than the parallel ones. In Fig.~\ref{fig:test_26_series} it can be noticed that around propagation no.~125 a substantial backtracking phase takes place. Such an event leads the domains and valid tuples in the table to increase again, advantaging $CT^{uf}_{CU}$ over the serial propagator.

\subsection{\texorpdfstring{$CT^{uf}_{CU}$ and Gecode}{CT-CU-uf and Gecode}}
\label{sub:gecode}
We now present some of the results obtained by comparing  $CT^{uf}_{CU}$ against 
Gecode (version 6.3.0). \enr{Both solvers rely on a single thread (host side) with no parallelization features enabled.}
The comparison involves the second set of instances of Section~\ref{sub:testGen}.
The instances have a number of variables between 100 and 150, a maximum domain size of 2000, and between 10000 and 15000 tuples in the table. The number of propagations for the tests in Table~\ref{tbl:gecode} ranges between 35000 and 85000. On some instances we noticed a slightly different number of propagations in favor of {\sc MiniCP}. This difference is less than $2\%$ and is likely due to $CT^{uf}_{CU}$ relying on the built-in method $\mathit{changed}()$ and not using \emph{lastSizes} to update $s^{val}$. However, the number of explored nodes and failures is always the same for both solvers. Note that while the instance features (such as number of tuples or domain sizes) are comparable with those of $LIN\_EB$ instances, the number of propagations (see also the tabular data in the Appendix) result very different. This is due to the fact that all variables are now involved in the table constraint.
Table~\ref{tbl:gecode} reports an excerpt of the instances we used, along with the solve times obtained by Gecode and $CT^{uf}_{CU}$.

\begin{table}[tb]
  {  \small
    \centering
    \begin{tabular*}{0.4\textwidth}{@{\extracolsep{\fill}}crrr}
	    {\multirow{2}{*}{Instance}} & \multicolumn{2}{c}{Solve times (s)} &  {\multirow{2}{*}{Speedup}} \\
        \cmidrule{2-3}
                          & Gecode & $CT^{uf}_{CU}$ &  \\        
       \midrule
        Ge\_1 & 37.2  & 23.9 & 1.55 \\ 
        Ge\_2 & 27.6  & 15.4 & 1.79 \\ 
        Ge\_3 & 33.6  & 16.6 & 2.02 \\ 
        Ge\_4 & 35.6  & 16.7 & 2.13 \\ 
        Ge\_5 & 42.3  & 20.3 & 2.07 \\ 
        Ge\_6 & 30.7  & 15.7 & 1.95 \\ 
        Ge\_7 & 31.3  & 17.9 & 1.74 \\ 
        Ge\_8 & 40.2  & 17.1 & 2.35 \\ 
        Ge\_9 & 37.6  & 20.9 & 1.80 \\ 
        Ge\_10 & 34.3 & 21.9 & 1.57 
    \end{tabular*}
    \hspace*{10mm}
    \begin{tabular*}{0.4\textwidth}{@{\extracolsep{\fill}}crrr}
	    {\multirow{2}{*}{Instance}} & \multicolumn{2}{c}{Solve times (s)} &  {\multirow{2}{*}{Speedup}} \\
        \cmidrule{2-3}
                          & Gecode & $CT^{uf}_{CU}$ &  \\        
       \midrule
        Ge\_11 & 35.0 & 17.1 & 2.04 \\ 
        Ge\_12 & 44.3 & 27.1 & 1.63 \\ 
        Ge\_13 & 29.1 & 13.7 & 2.13 \\ 
        Ge\_14 & 45.2 & 20.0 & 2.26 \\ 
        Ge\_15 & 28.5 & 14.9 & 1.91 \\ 
        Ge\_16 & 27.2 & 14.6 & 1.87 \\ 
        Ge\_17 & 36.3 & 16.1 & 2.25 \\ 
        Ge\_18 & 28.9 & 19.3 & 1.50 \\ 
        Ge\_19 & 26.5 & 14.8 & 1.80 \\ 
        Ge\_20 & 34.1 & 18.0 & 1.88 \\ 
    \end{tabular*}
	\caption{The solving times (in seconds) for Gecode and $CT^{uf}_{CU}$.} 
    \label{tbl:gecode}}
\end{table}

\section{Conclusions}
\label{sec:conclusion}
In this paper we presented a GPU parallel version of the propagation algorithm for the table constraint,
a fundamental built-in constraint for Constraint Logic Programming languages.
The massive parallelism offered by GPU computing allows such implementations to scale up as the CSP instance size grows.  
Such scaling was especially noticeable in the case of CT\mbox{{}$_{CU}^{uf}$} (Section~\ref{sub:tests}).
We have experimentally observed the speedup provided by the CUDA parallel implementations when compared to the serial counterpart.
Such implementations, in particular CT\mbox{{}$_{CU}^{uf}$}, have been shown to be capable of achieving a significant speedup also against a state-of-the-art solver such as Gecode which relies on more sophisticated propagation techniques.

\bibliographystyle{acmtrans}
\bibliography{iclp25Table}

\appendix

\begin{table}[b]
    \small
    \centering
    \begin{tabular*}{0.8\textwidth}{@{\extracolsep{\fill}}crrrrrr}
	    {\multirow{2}{*}{Instance}} & \multicolumn{3}{c}{Solve times (sec)} & \multicolumn{2}{c}{Speedup} & {\multirow{2}{*}{Propagations ($10^6$)}} \\
        \cmidrule{2-4}\cmidrule{5-6}
                          & Serial & $CT^{f}_{CU}$ & $CT^{uf}_{CU}$ & $CT^{f}_{CU}$  & $CT^{uf}_{CU}$ &  \\
        \midrule
            LIN\_B\_1  & 344.6 & 147.6 & 111.4 & 2.33 & 3.09 & 3.72 \\ 
            LIN\_B\_2  & 119.1 & 58.4 & 46.7 & 2.04 & 2.55 & 2.02 \\ 
            LIN\_B\_3  & T.O. & T.O. & T.O. & - & -  & - \\ 
            LIN\_B\_4  & 80.5 & 32.1 & 21.7 & 2.51 & 3.71 & 0.62 \\ 
            LIN\_B\_5  & 181.4 & 125.7 & 87.2 & 1.44 & 2.08 & 3.43 \\ 
            LIN\_B\_6  & T.O. & T.O. & T.O. & - & - & - \\ 
            LIN\_B\_7  & 150.8 & 86.0 & 64.5 & 1.75 & 2.34 & 2.59 \\ 
            LIN\_B\_8  & 24.5 & 5.2 &   2.8 & 4.70 & 8.61 & 0.07 \\ 
            LIN\_B\_9  & 35.5 & 15.2 & 10.2 & 2.34 & 3.47 & 0.33 \\ 
            LIN\_B\_10 & 145.5 & 108.1 & 83.2 & 1.35 & 1.75  & 2.99 \\ 
            LIN\_B\_11 & 181.3 & 109.0 & 83.2 & 1.66 & 2.18 & 3.15 \\ 
            LIN\_B\_12 & 256.9 & 151.2 & 121.4 & 1.70 & 2.12 & 4.53 \\ 
            LIN\_B\_13 & 140.7 & 104.6 & 80.1 & 1.34 & 1.76 & 3.33 \\ 
            LIN\_B\_14 & 192.5 & 105.7 & 84.4 & 1.82 & 2.28  & 2.77 \\ 
            LIN\_B\_15 & T.O. & T.O. & T.O. & - & - & - \\ 
            LIN\_B\_16 & 385.7 & 168.8 & 135.9 & 2.28 & 2.84 & 4.71 \\ 
            LIN\_B\_17 & 290.4 & 156.9 & 125.3 & 1.85 & 2.32  & 4.72 \\ 
            LIN\_B\_18 & 219.6 & 119.2 & 94.2 & 1.84 & 2.33 & 2.89 \\ 
            LIN\_B\_19 & 342.9 & 156.8 & 124.0 & 2.19 & 2.76 &4.00 \\ 
            LIN\_B\_20 & 32.7 & 12.5 & 7.7 & 2.61 & 4.24  & 0.22 \\ 
            LIN\_B\_21 & 269.8 & 112.0 & 84.0 & 2.41 & 3.21 & 2.61  \\ 
            LIN\_B\_22 & 342.5 & 139.2 & 104.1 & 2.46 & 3.29 & 3.32 \\ 
            LIN\_B\_23 & 160.6 & 72.7 & 55.5 & 2.21 & 2.89  & 1.58 \\ 
            LIN\_B\_24 & 204.9 & 146.8 & 107.1 & 1.40 & 1.91   & 4.27 \\ 
            LIN\_B\_25 & 168.9 & 98.9 & 75.5 & 1.71 & 2.24 & 2.73 \\ 
            LIN\_B\_26 & 370.9 & 142.1 & 111.6 & 2.61 & 3.32 & 3.36 \\ 
            LIN\_B\_27 & 362.8 & 147.0 & 120.6 & 2.47 & 3.01 & 3.58 \\ 
            LIN\_B\_28 & 220.1 & 130.7 & 110.5 & 1.68 & 1.99  & 4.40 \\ 
            LIN\_B\_29 & 380.5 & 145.4 & 113.0 & 2.62 & 3.37 & 3.28 \\ 
            LIN\_B\_30 & 229.6 & 133.0 & 108.3 & 1.73 & 2.12 & 3.21 \\ 
    \end{tabular*}
	\caption{MiniZinc solve times, number of propagations and speedup for $CT^{f}_{CU}$ and $CT^{uf}_{CU}$ on the first batch of tests. A barplot summing up the table is presented in Fig.~\ref{fig:sumUp4090_b}. 
    }
    \label{tbl:lin_b}
\end{table}

\begin{table}[tb]
      \small
  \centering
    \begin{tabular*}{0.8\textwidth}{@{\extracolsep{\fill}}crrrrrr@{}}
	    \multirow{2}{*}{Instance} & \multicolumn{3}{c}{Solve times (sec)} & \multicolumn{2}{c}{Speedup} & \multirow{2}{*}{Propagations ($10^6$)} \\
        \cmidrule{2-4}\cmidrule{5-6}
                          & Serial & $CT^{f}_{CU}$ & $CT^{uf}_{CU}$ & $CT^{f}_{CU}$  & $CT^{uf}_{CU}$ &  \\
        \midrule
        LIN\_EB\_1  & 681.5 & 347.2 & 231.7 & 1.96 & 2.94 & 6.84 \\ 
        LIN\_EB\_2  & 429.7 & 208.3 & 153.9 & 2.06 & 2.79 & 3.96 \\ 
        LIN\_EB\_3  & T.O. & T.O. & T.O. & - & - & -  \\ 
        LIN\_EB\_4  & 157.1 & 26.5 & 9.0 & 5.93 & 17.44 & 0.08 \\ 
        LIN\_EB\_5  & 226.8 & 86.3 & 57.7 & 2.63 & 3.93 & 1.87 \\
        LIN\_EB\_6  & 243.0 & 55.3 & 25.9 & 4.39 & 9.38 & 0.48 \\ 
        LIN\_EB\_7  & T.O. & T.O. & T.O. & - & - & - \\ 
        LIN\_EB\_8  & 300.1 & 220.5 & 154.3 & 1.36 & 1.94 & 5.56  \\ 
        LIN\_EB\_9  & 410.2 & 253.9 & 196.7 & 1.62 & 2.09 & 6.09  \\ 
        LIN\_EB\_10 & 656.6 & 439.7 & 313.8 & 1.49 & 2.09 & 13.22 \\ 
        LIN\_EB\_11 & T.O. & T.O. & T.O. & - & - & -  \\ 
        LIN\_EB\_12 & 293.6 & 223.6 & 160.8 & 1.31 & 1.83 & 5.68  \\ 
        LIN\_EB\_13 & 427.8 & 276.6 & 194.6 & 1.55 & 2.20 & 5.85  \\ 
        LIN\_EB\_14 & 13.3 & 4.6 & 3.0 & 2.89 & 4.46 & 0.07 \\ 
        LIN\_EB\_15 & T.O. & 578.1 & 375.4 & $ > $ 1.56 & $ > $ 2.40 & 14.04    \\ 
        LIN\_EB\_16 & T.O. & T.O. & 744.2 & - & $ > $ 1.21 & 13.19 \\ 
        LIN\_EB\_17 & T.O. & 366.4 & 262.2 & $ > $ 2.46 & $ > $ 3.43 & 8.92   \\
        LIN\_EB\_18 & 12.2 & 3.9 & 2.5 & 3.10 & 4.81 & 0.06 \\ 
        LIN\_EB\_19 & 617.6 & 348.4 & 262.3 & 1.77 & 2.35 & 6.97  \\ 
        LIN\_EB\_20 & 168.8 & 29.0 & 19.5 & 5.82 & 8.67 & 0.28 \\ 
        LIN\_EB\_21 & T.O. & T.O. & T.O. & - & - & - \\ 
        LIN\_EB\_22 & T.O. & 405.9 & 286.8 & $ > $2.22 & $ > $3.14 & 6.45   \\ 
        LIN\_EB\_23 & 16.6 & 5.2 & 3.4 & 3.20 & 4.95 & 0.07   \\ 
        LIN\_EB\_24 & T.O. & 513.3 & 328.8 & $ > $1.75 & $ > $2.74 & 10.33  \\ 
        LIN\_EB\_25 & T.O. & T.O. & T.O. & - & - & -  \\ 
        LIN\_EB\_26 & 84.1 & 24.5 & 6.7 & 3.40 & 12.53 & 0.13 \\ 
        LIN\_EB\_27 & 837.5 & 440.6 & 289.8 & 1.90 & 2.89 & 7.34 \\ 
        LIN\_EB\_28 & 545.7 & 290.2 & 192.0 & 1.88 & 2.84 & 5.57  \\ 
        LIN\_EB\_29 & 12.0 & 4.7 & 3.4 & 2.62 & 3.59 & 0.08 \\ 
        LIN\_EB\_30 & 366.1 & 241.2 & 171.0 & 1.52 & 2.14 & 4.84  \\ 
    \end{tabular*}
    \caption{MiniZinc solve times, number of propagations and speedup for $CT^{f}_{CU}$ and $CT^{uf}_{CU}$ on the second batch of tests. A barplot summing up the table is presented in Fig.~\ref{fig:sumUp4090_eb}.   
    }
    \label{tbl:lin_eb}
\end{table}

\section{Results of the experiments}

In this section we report the tables summarizing the performance of the tested implementations across $LIN\_B$ and $LIN\_EB$ batches. Such results are summarized in Figures~\ref{fig:sumUp4090_b} and~\ref{fig:sumUp4090_eb} and included here in detail for completeness.

Solve times entries in Tables~\ref{tbl:lin_b} and~\ref{tbl:lin_eb} presenting a \emph{T.O.} value indicate that such a test timed out after 15 minutes. 

Let $CT_{CU}$ denote either $CT_{CU}^{f}$ or $CT_{CU}^{uf}$, its relative speedup is computed as $\frac{serial\;sovle\;time }{CT_{CU} \;solve \;time}$.
Table~\ref{tbl:lin_eb} presents some instances where the serial solver timed out but $CT_{CU}$ did not. In such scenarios a lower bound on the achievable speedup is reported, calculated by imposing \emph{serial solve time} to the 15 minute lower bound, i.e., $\frac{900}{CT_{CU} \;solve\;time}$.

\end{document}